\documentclass[journal,twoside]{IEEEtran}
\ifCLASSINFOpdf
\else
   \usepackage[dvips]{graphicx}
\fi
\usepackage[dvipsnames]{xcolor}  % -J  might clash option, need to be defined earlier
\usepackage{url}
\usepackage{graphicx}
\usepackage{amsmath,amssymb} % define this before the line numbering.
\usepackage[sort]{cite}  % -J
\usepackage{hyperref}  % -J if 'Paragraph ended before' error --> clear .aux files
\usepackage[all]{hypcap}  % -J

\hypersetup{
  linkcolor  = MidnightBlue,
  citecolor  = MidnightBlue,
  urlcolor   = MidnightBlue,
  colorlinks = true,
}

\begin{document}
\setlength{\parskip}{0pt}  % -J
\bstctlcite{IEEEexample:BSTcontrol}
\title{S3: A Spectral-Spatial Structure Loss for Pan-Sharpening Networks}
\author{
Jae-Seok Choi, Yongwoo Kim, and Munchurl Kim, \IEEEmembership{Senior Member, IEEE}
\thanks{This paper was submitted for review on Apr. 9, 2019 \textit{(Corresponding author: Munchurl Kim)}. This research was supported by Basic Science Research Program through the National Research Foundation of Korea (NRF) funded by the Ministry of Science, ICT \& Future Planning (No. 2017R1A2A2A05001476). We thank Korea Aerospace Research Institute for providing the KOMPSAT-3A satellite dataset for our experiments.}
\thanks{J.-S. Choi and M. Kim are with the School of Electrical Engineering, Korea Advanced Institute of Science and Technology, Daejeon 34141, South Korea (e-mail: jschoi14@kaist.ac.kr; mkim@ee.kaist.ac.kr). Y. Kim is with the Artificial Intelligence Research Division, Korea Aerospace Research Institute, Daejeon 34133, South Korea (e-mail: ywkim85@kari.re.kr).}
}
%\thanks{This paragraph of the first footnote will contain the date on which you submitted your paper for review. It will also contain support information, including sponsor and financial support acknowledgment. For example, ``This work was supported in part by the U.S. Department of Commerce under Grant BS123456.'' }
%\thanks{The next few paragraphs should contain the authors' current affiliations, including current address and e-mail. For example, F. A. Author is with the National Institute of Standards and Technology, Boulder, CO 80305 USA (e-mail: author@boulder.nist.gov).}
%\markboth{IEEE Geoscience and Remote Sensing Letters, Vol. 99, No. 99, Mar. 2019}
%{Choi \MakeLowercase{\textit{et al.}}: S3: A Spectral-Spatial Structure Loss for Pan-Sharpening Networks}
\maketitle
\begin{abstract}
Recently, many deep-learning-based pan-sharpening methods have been proposed for generating high-quality pan-sharpened (PS) satellite images. These methods focused on various types of convolutional neural network (CNN) structures, which were trained by simply minimizing a spectral loss between network outputs and the corresponding high-resolution multi-spectral (MS) target images. However, due to different sensor characteristics and acquisition times, high-resolution panchromatic (PAN) and low-resolution MS image pairs tend to have large pixel misalignments, especially for moving objects in the images. Conventional CNNs trained with only the spectral loss with these satellite image datasets often produce PS images of low visual quality including double-edge artifacts along strong edges and ghosting artifacts on moving objects. In this letter, we propose a novel loss function, called a spectral-spatial structure (S3) loss, based on the correlation maps between MS targets and PAN inputs. Our proposed S3 loss can be very effectively utilized for pan-sharpening with various types of CNN structures, resulting in significant visual improvements on PS images with suppressed artifacts.
\end{abstract}


\IEEEpeerreviewmaketitle

\section{Introduction}
\IEEEPARstart{D}{ue} to their sensor resolution constraints and bandwidth limitation, satellites often acquire multi-resolution multi-spectral images of the same target areas. In general, satellite images include pairs of a low-resolution (LR) multi-spectral image (MS) of longer ground sample distance (GSD), and a high-resolution (HR) panchromatic (PAN) image of shorter GSD. By extracting high-quality spatial structures from a PAN image and multi-spectral information from an MS image, one can generate a pan-sharpened (PS) image which has the same GSD as that of the PAN image but with the spectral information of the MS image. This is known as pan-sharpening or pan-colorization.

\subsection{Related Works}
Traditional pan-sharpening methods \cite{Gillespie1986,J.Carper1990,Shah2008,Kang2014,Xu2014,mallat1989theory,starck2007undecimated,Pan2013,He2014,Palsson2014,Vivone2015,Garzelli2008,Brodu2017,Alparone2008}  include component substitution \cite{Gillespie1986,J.Carper1990,Shah2008,Kang2014,Xu2014}, multiresolution analysis \cite{mallat1989theory,starck2007undecimated} and machine-learning \cite{Pan2013,He2014,Palsson2014}. Comparisons for component substitution and multiresolution analysis based approaches were presented thoroughly in \cite{Vivone2015}. Component substitution based methods often incorporated the Brovey transform (BT) \cite{Gillespie1986}, the intensity-hue-saturation \cite{J.Carper1990}, principal component analysis (PCA) \cite{Shah2008}, or matting models \cite{Kang2014} for pan-sharpening. In multiresolution analysis based methods, the spatial structures of PAN images are decomposed using wavelet \cite{mallat1989theory} or undecimated wavelet \cite{starck2007undecimated} decomposition techniques, and are fused with up-sampled MS images to produce PS images. These methods have relatively low computation complexity but tend to produce PS images with mismatched spectral information. Machine-learning based methods \cite{Pan2013,He2014,Palsson2014} learn pan-sharpening models by optimizing a loss function of inputs and targets with some regularization terms. \par
With the advent of deep-learning, recent pan-sharpening methods \cite{Masi2016,Yang2017,Huang2015,Scarpa2018,Lanaras2018,zhang2019pan} started to incorporate various types of convolutional neural network (CNN) structures and are showing a large margin of quality improvements over traditional pan-sharpening methods. Most of these CNN-based pan-sharpening methods utilized network structures that were proven to be effective in classification \cite{He2015,huang2017densely} and super-resolution (SR) \cite{dong2014learning,kim2016accurate,lim2017enhanced} tasks. As the goal for pan-sharpening is to increase the resolution of MS inputs, many conventional CNN-based pan-sharpening methods employed network structures from the previous CNN-based SR methods \cite{dong2014learning,kim2016accurate,lim2017enhanced}. Pan-sharpening CNN (PNN) \cite{Masi2016} is known as the first method to employ CNN into pan-sharpening. The PNN used a shallow 3-layered network adopted from SRCNN \cite{dong2014learning} which is the first CNN-based SR method. The PNN was trained and tested on the Ikonos, GeoEye-1 and WorldView-2 satellite image datasets. Inspired by the success of ResNet \cite{He2015} in classification, PanNet \cite{Yang2017} incorporated the ResNet architecture with a smaller number of filter parameters to perform pan-sharpening. Lanaras \textit{et al.} \cite{Lanaras2018} employed the state-of-the-art SR network, EDSR \cite{lim2017enhanced}, and proposed a moderately deep network version (DSen2) and a very deep network version (VDSen2) for pan-sharpening. Recently, a bidirectional pyramid network (BDPN) \cite{zhang2019pan} has been proposed, using deep and shallow networks for PAN and MS inputs separately. 

\subsection{Our Contributions}
Since the state-of-the-art CNN-based pan-sharpening methods, PanNet \cite{Yang2017}, DSen2 \cite{Lanaras2018} and BDPN \cite{zhang2019pan} were trained using a simple spectral loss function for minimizing reconstruction error between generated images and MS target images, their PS result images often suffer from visually unpleasant artifacts along building edges and on moving cars in their resulting PS images of shorter GSD such as the WorldView-3 dataset. This is because, as GSD becomes smaller, pixel misalignments between PAN and MS inputs tend to get larger due to inevitable acquisition time difference and mosaicked sensor arrays. In such scenarios, the spectral loss between network outputs and MS target images are insufficient for training, thus resulting in the PS images of low visual quality.

\begin{figure}[t]
\centering
\includegraphics[width=0.47 \textwidth]{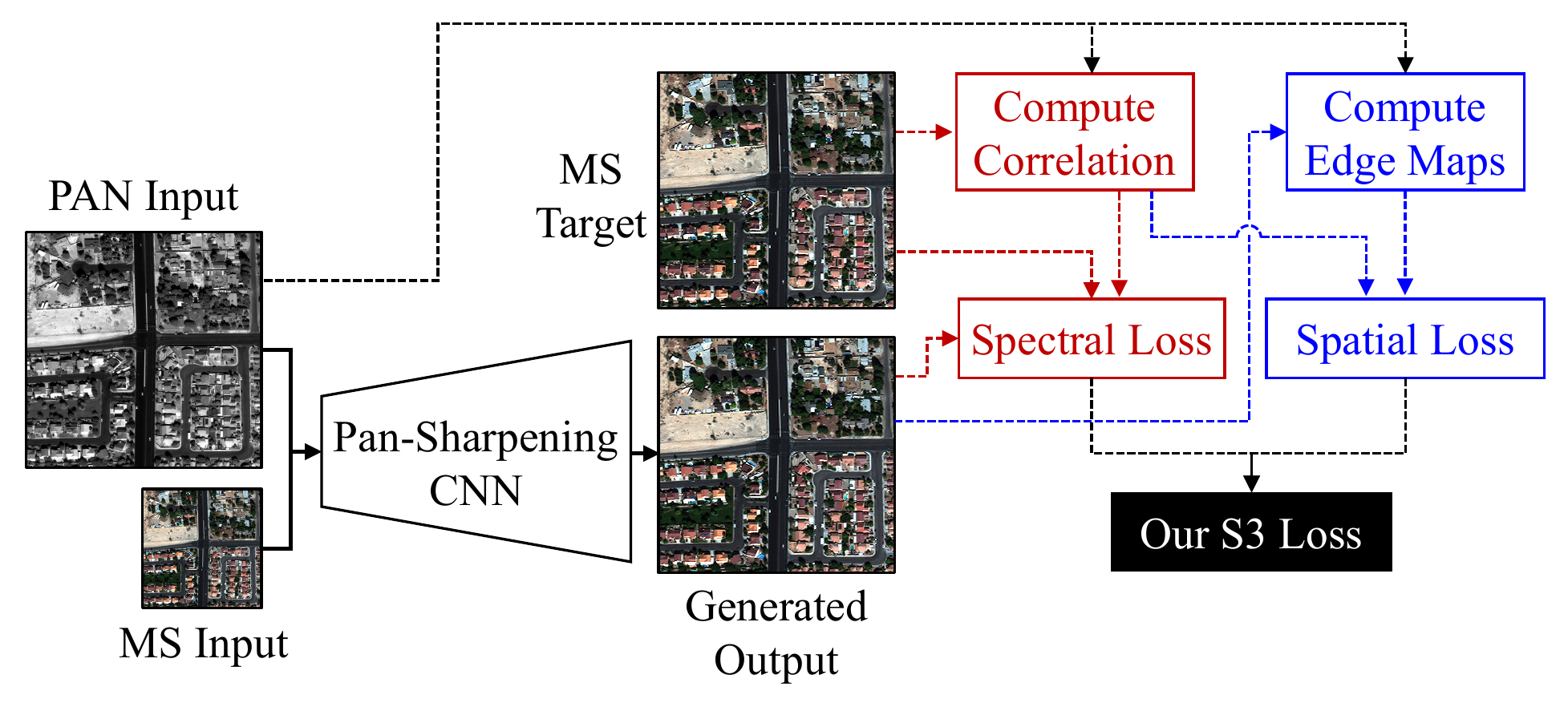}
\caption{A framework for CNN-based pan-sharpening with our proposed spectral-spatial structure (S3) loss.}
\label{fig:1}
\end{figure}

In this letter, we propose a novel loss term, called a spectral-spatial structure (S3) loss, which can be effectively utilized for training of pan-sharpening CNNs to learn spectral information of MS targets while preserving the spatial structure of PAN inputs. Our S3 loss consists of two loss functions: a spectral loss between network outputs and MS targets, and a spatial loss between network outputs and PAN inputs. Here, both spectral and spatial losses are computed based on the correlation maps between MS targets and PAN inputs. The spectral loss is selectively applied for the areas where averaged MS targets and PAN inputs are highly correlated. The spatial loss only considers gradient maps of generated images (network output) and PAN inputs. In doing so, our network using the S3 loss can generate PS images where double-edge artifacts and ghosting artifacts on moving cars are significantly reduced. Finally, we show that our S3 loss can effectively work with various pan-sharpening CNNs. Fig. \ref{fig:1} shows a CNN-based pan-sharpening architecture with our proposed S3 loss.

\section{Proposed Method}
\subsection{Formulations}
Most of satellite imagery datasets include PAN images of higher resolution (smaller GSD) $\mathbf{P}_0$, and the corresponding MS images of lower resolution (larger GSD) $\mathbf{M}_1$. Here, the subscripts of $\mathbf{P}_0$ and $\mathbf{M}_1$ denote a level of resolution where a smaller number is for a higher resolution. We have two scenarios in terms of scales (resolutions): (i) Our final goal in pan-sharpening is to utilize both $\mathbf{P}_0$ and $\mathbf{M}_1$ inputs to generate a high-quality PS image $\mathbf{G}_0$, which has the same resolution as $\mathbf{P}_0$, while preserving spectral information of $\mathbf{M}_1$. This case corresponds to the \textit{original scale} scenario in \cite{Yang2017,Lanaras2018}; (ii) Now we consider a pan-sharpening model that requires training using input and target pairs. For target images, we use $\mathbf{M}_1$. For input images, we use $\mathbf{M}_2$ and $\mathbf{P}_1$, which are down-scaled versions of $\mathbf{M}_1$ and $\mathbf{P}_0$ respectively, using a degradation model \cite{Lanaras2018}. The pan-sharpening CNN takes $\mathbf{M}_2$ and $\mathbf{P}_1$ as inputs, and generates $\mathbf{G}_1$. This case corresponds to the \textit{lower scale} scenario in \cite{Yang2017,Lanaras2018}. In conclusion, training and testing the pan-sharpening networks are performed under the \textit{lower} and \textit{original scale} scenarios, respectively. In this regard, the conventional pan-sharpening networks were trained by simply minimizing a spectral loss between network outputs $\mathbf{G}_1$ and MS targets $\mathbf{M}_1$ under the \textit{lower scale} scenario.
\subsection{Proposed S3 Loss}
We now define our spectral-spatial structure (S3) loss, which can be used for training any pan-sharpening CNN to yield high-quality PS images $\mathbf{G}_1$, and ultimately $\mathbf{G}_0$. First, we define our feedforward pan-sharpening operation as
\begin{align}  
	\mathbf{G}_1 = g(\mathbf{M}_2,\mathbf{P}_1,\theta),
\label{eq:1}
\end{align}
where $g$ is a pan-sharpening CNN with filter parameters $\theta$. The conventional methods \cite{Yang2017,Lanaras2018} use the L2 loss as
\begin{align}  
	\theta^\ast = \operatorname*{argmin}_\theta \sum{\lVert g(\mathbf{M}_2,\mathbf{P}_1,\theta) - \mathbf{M}_1 \rVert_2^2}.
\label{eq:2}
\end{align}
However, solely using this loss function for training often leads to artifacts in resultant images $\mathbf{G}_1$, due to inherent misalignments between $\mathbf{M}_1$ and $\mathbf{P}_1$. To overcome this limitation, we propose S3 loss consisting of two loss functions: a spectral loss between $\mathbf{G}_1$ and $\mathbf{M}_1$; and a spatial loss between $\mathbf{G}_1$ and $\mathbf{P}_1$. First, the spectral loss more penalizes the spectral distortion on the areas where grayed $\mathbf{M}_1$ (denoted as $\hat{\mathbf{M}}_1$) and $\mathbf{P}_1$ are highly correlated. The correlation map $\mathbf{S}$ can be formulated as
\begingroup  % -J
\allowdisplaybreaks  % -J
\begin{alignat}{5}
cov(\hat{\mathbf{M}}_1,\mathbf{P}_1)&=m(\hat{\mathbf{M}}_1\odot\mathbf{P}_1)-m(\hat{\mathbf{M}}_1)\odot m(\mathbf{P}_1),
\label{eq:3} \\
std(\hat{\mathbf{M}}_1)&=\sqrt{|cov(\hat{\mathbf{M}}_1,\hat{\mathbf{M}}_1)|+e},
\label{eq:4} \\
std(\mathbf{P}_1)&=\sqrt{|cov(\mathbf{P}_1,\mathbf{P}_1)|+e},
\label{eq:5} \\
corr(\hat{\mathbf{M}}_1,\mathbf{P}_1)&=\frac{cov(\hat{\mathbf{M}}_1,\mathbf{P}_1)}{std(\hat{\mathbf{M}}_1) \odot std(\mathbf{P}_1)},
\label{eq:6} \\
\mathbf{S}&=|corr(\hat{\mathbf{M}}_1,\mathbf{P}_1)|^\gamma,
\label{eq:7}
\end{alignat}
\endgroup  % -J
where $m$ is a mean filter, $\odot$ denotes an element-wise multiplication, $\gamma$ is a control parameter, and $e$ is a very small value, i.e. $10^{-10}$. We use a 31$\times$31 box filter for $m$. We empirically set $\gamma$ to 4. Using $\mathbf{S}$, our spectral loss $L_c$ is then defined as
\begin{align}
	L_c = \sum{\lVert (\mathbf{G}_1 - \mathbf{M}_1) \odot \mathbf{S} \rVert_1^1}.
\label{eq:8}
\end{align}
Here, we try to minimize spectral loss between $\mathbf{G}_1$ and $\mathbf{M}_1$ only for pixel areas where $\hat{\mathbf{M}}_1$ and $\mathbf{P}_1$ have large positive and negative correlations. Note that $\mathbf{S}$ is not trainable.

For our spatial loss $L_a$, we try to minimize the difference between the gradient map of grayed $\mathbf{G}_1$ (denoted as $\hat{\mathbf{G}}_1$) and that of $\mathbf{P}_1$, which is formulated as
\begin{align}  
	L_a = \sum{\lVert (grad(\hat{\mathbf{G}}_1)-grad(\mathbf{P}_1)) \odot (2-\mathbf{S}) \rVert_1^1},
\label{eq:9}
\end{align}
where $grad$ for $\mathbf{X}$ is a function defined as
\begin{align}  
	grad(\mathbf{X})=\frac{\mathbf{X}-m(\mathbf{X})}{std(\mathbf{X})}.
\label{eq:10}
\end{align}
We incorporated $(2-\mathbf{S})$ into $L_a$ in (\ref{eq:9}), so that $L_a$ focuses more on those areas where $L_c$ is less focused. Finally, combining $L_c$ and $L_a$, we have our final S3 loss $L_{S3}$ as
\begin{align}  
	L_{S3}=L_c+w_a L_a,
\label{eq:11}
\end{align}
where $w_a$ is a weighting value. We empirically set $w_a$ to 1.

In order to show the effectiveness of our S3 loss, we incorporated our S3 loss into the state-of-the-art pan-sharpening networks, PanNet \cite{Yang2017}, BDPN \cite{zhang2019pan} and DSen2 \cite{Lanaras2018}, which are named as PanNet-S3, DSen2-S3 and DSen2-S3 in our experiments. The DSen2 network has 14 convolutional layers with 128 channels, having about 1.8M filter parameters. PanNet has 10 layers with 76K parameters, while BDPN has 46 layers with 1.4M parameters. As for our PanNet-S3 and BDPN-S3, full data of MS-PAN inputs were concatenated and used as the network input.

\begin{figure*}[t]
\centering
\includegraphics[width=1 \textwidth]{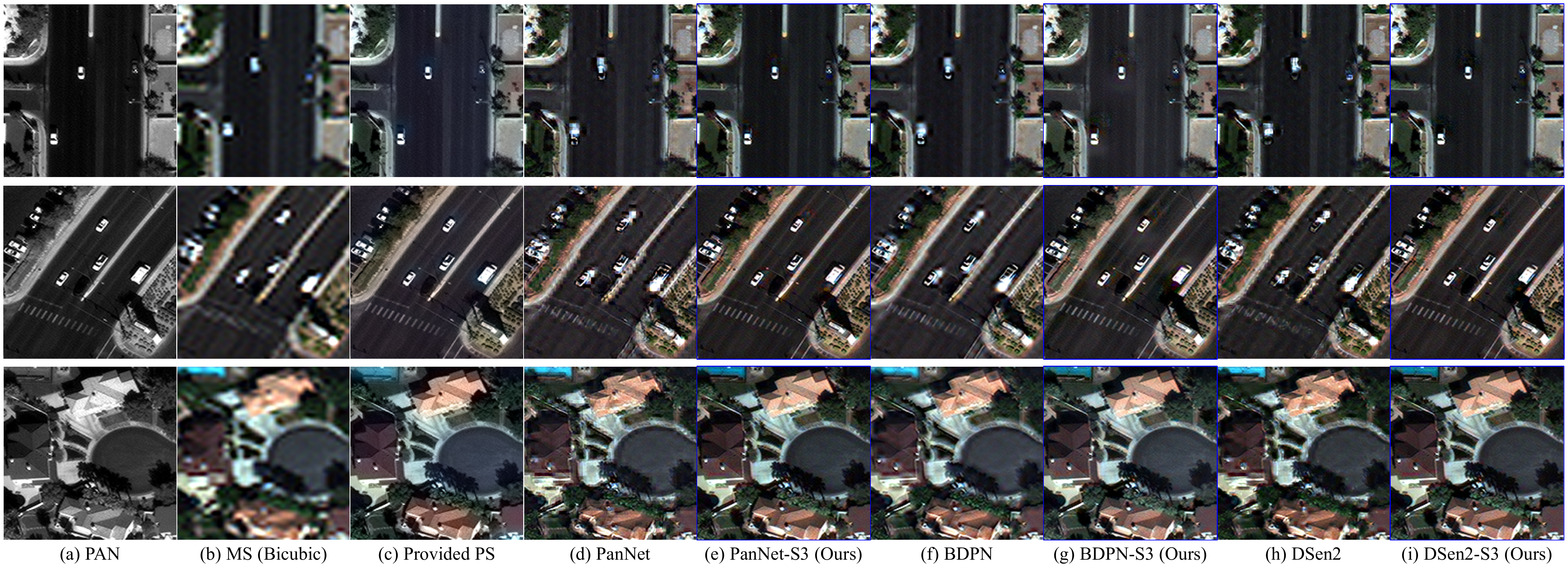}
\caption{Pan-sharpening results at the original scale using various methods on the WorldView-3 dataset (best viewed when zoomed in).}
\label{fig:2}
\end{figure*}

\begin{figure}[t]
\centering
\includegraphics[width=240pt]{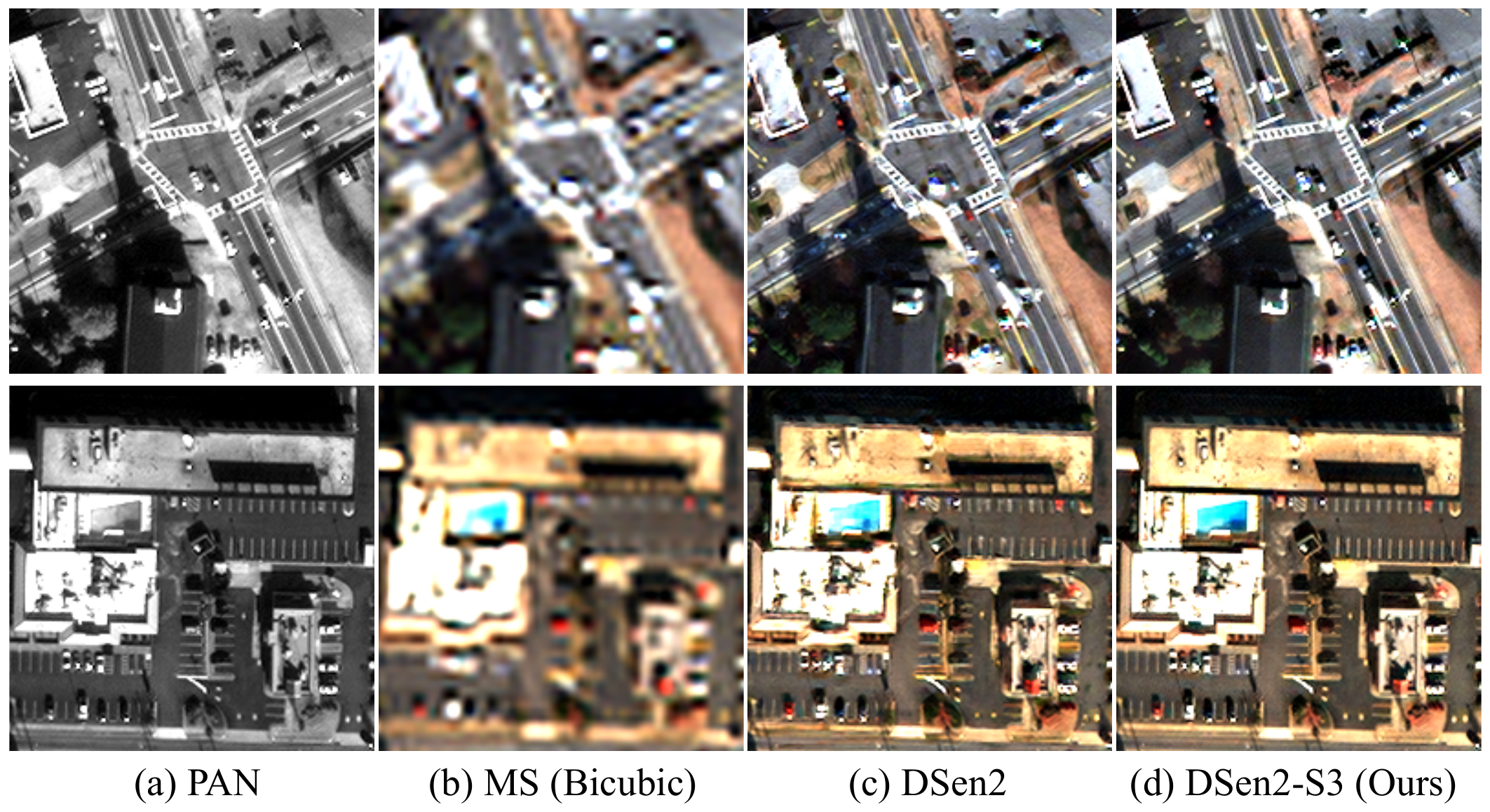}
\caption{Additional pan-sharpening results on the WorldView-2 dataset.}
\label{fig:3}
\end{figure}

\begin{figure}[t]
\centering
\includegraphics[width=240pt]{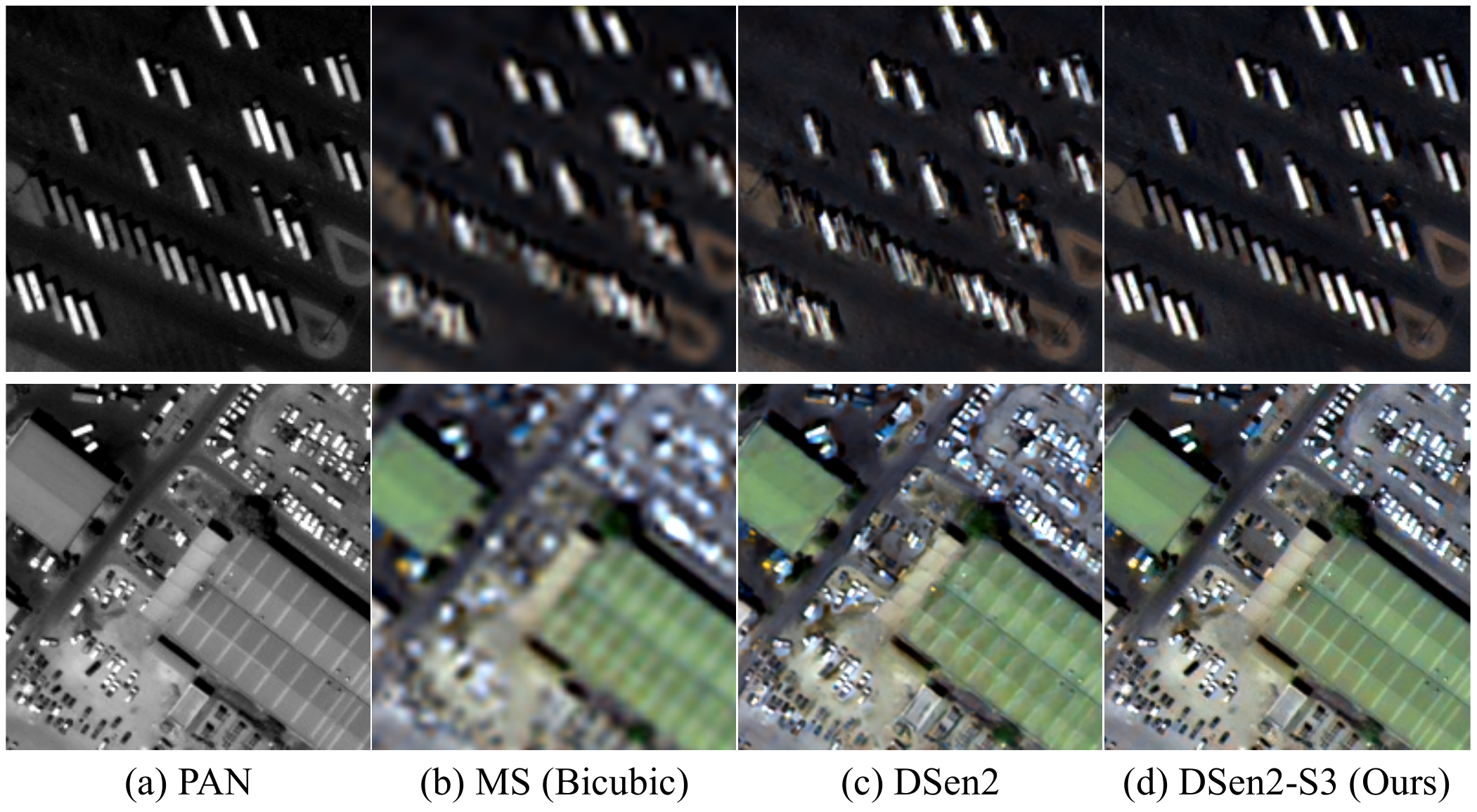}
\caption{Additional pan-sharpening results on the KOMPSAT-3A dataset.}
\label{fig:4}
\end{figure}

\begin{figure}[t]
\centering
\includegraphics[width=180pt]{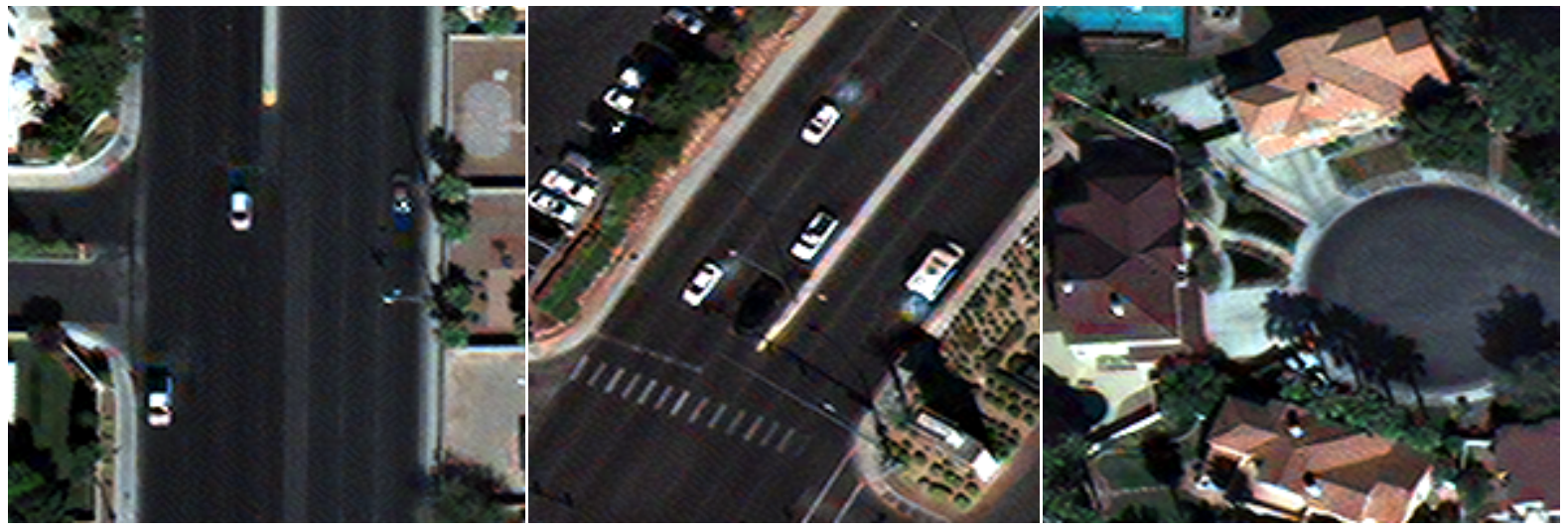}
\caption{Pan-sharpening results for our DSen2-S3 without using the correlation map $\mathbf{S}$.}
\label{fig:5}
\end{figure}

\section{Experiment Results and Discussions}
\subsection{Experiment Settings}
\subsubsection{Datasets}
All the networks including ours and baselines were trained and tested on the WorldView-3 satellite image dataset, whose PAN images are of about 0.3 $m$ GSD and MS images are of about 1.2 $m$ GSD. PS images of 0.3 $m$ GSD are also provided in the dataset, but they are used only for a visual comparison purpose with our results. Note that the WorldView-3 satellite image dataset has the shortest GSD (highest-resolution) among aforementioned datasets. We selected and used the WorldView-3 satellite image dataset from SpaceNet Challenge dataset \cite{spaceneturl}. The RGB channels of the MS images were used for all experiments. Total 13K MS-PAN image pairs were used for training networks, where cropping and various data augmentations were conducted on the fly during the training. The MS-PAN training subimages were created by applying a down-scaling method in \cite{Lanaras2018}. The cropped MS subimages used for training are 32$\times$32-sized, while PAN subimages are of 128$\times$128 size. Before being fed into the networks, the training image pairs were normalized to have a range between 0 and 1. Training was done in the \textit{lower scale} scenario.

\begin{table}[b]
\centering
\caption{Average quality metric scores for 100 result images at the original scale on the WorldView-3 test dataset.}
\label{table:1}
\renewcommand{\arraystretch}{1.1}% Spread rows out...
\scalebox{0.83}{
\setlength\tabcolsep{3pt} % default value: 6pt
\begin{tabular} {c|c|c|c|c}
\hline \hline
Method / Metric&Avg. ERGAS$_1$&Avg. SCC$_1$&Avg. SCC$_0$&Avg. n-ERGAS$_1$\\
\hline
Bicubic&0.6818$\pm$0.0401&0.8577$\pm$0.0089&0.4826$\pm$0.0126&0.6818$\pm$0.0401\\
\hline
Provided PS&3.6845$\pm$0.2036&\textbf{0.9647}$\pm$0.0014&\textbf{0.9679}$\pm$0.0012&3.3506$\pm$0.1849\\
\hline
PanNet \cite{Yang2017}&0.4360$\pm$0.0254&0.8468$\pm$0.0094&0.7367$\pm$0.0092&0.4360$\pm$0.0254\\
\hline
PanNet-S3 (Ours)&3.0647$\pm$0.1830&0.9530$\pm$0.0015&0.9568$\pm$0.0012&2.6465$\pm$0.1533\\
\hline
BDPN \cite{zhang2019pan}&1.3695$\pm$0.0779&0.8836$\pm$0.0069&0.9051$\pm$0.0039&1.3691$\pm$0.0779\\
\hline
BDPN-S3 (Ours)&3.3380$\pm$0.2043&0.9563$\pm$0.0015&0.9580$\pm$0.0012&2.8221$\pm$0.1681\\
\hline
DSen2 \cite{Lanaras2018}&\textbf{0.4278}$\pm$0.0258&0.8508$\pm$0.0088&0.6485$\pm$0.0142&\textbf{0.4278$\pm$0.0258}\\
\hline
DSen2-S3 (Ours)&3.1800$\pm$0.1929&0.9536$\pm$0.0015&0.9539$\pm$0.0013&2.6942$\pm$0.1575\\
\hline \hline
\end{tabular}}
\end{table}

\subsubsection{Training}
We trained all the networks using the decoupled ADAMW optimization \cite{loshchilov2017fixing} with an initial learning rate of $10^{-4}$, initial weight decay of $10^{-7}$, and the other hyper-parameters as defaults. The mini-batch size was set to 2. We employed a uniform weight initialization technique in \cite{jia2014caffe}. All the networks including our proposed networks were implemented using TensorFlow \cite{abadi2016tensorflow}, and were trained and tested on Nvidia Titan Xp GPU. The networks were trained for total $10^{6}$ iterations, where the learning rate and weight decay were lowered by a factor of 10 after $5\times10^{5}$ iterations. In our PanNet-S3, initial learning rate and weight decay were set to $5\times10^{-4}$ and $10^{-8}$, respectively. In our BDPN-S3, we used $10^{-8}$ for the hyper-parameter $e$ in our S3 loss, and $w_a$ in the S3 loss was empirically set to 2.

\subsection{Comparisons and Discussions}
We now compare our proposed methods using the S3 loss, with the conventional pan-sharpening methods including bicubic, PS images provided from the WorldView-3 dataset, PanNet \cite{Yang2017}, BDPN \cite{zhang2019pan} and DSen2 \cite{Lanaras2018}. We implemented PanNet, BDPN and DSen2 according to their technical descriptions, and trained them on the WorldView-3 dataset. At testing, for MS input images with a size of 160$\times$160, average computation time for our DSen2-S3 on GPU is about 2 sec per image. \par
As in \cite{Vivone2015,Yang2017}, we use two popular metrics: Erreur Relative Globale Adimensionnelle de Synthèse (ERGAS) \cite{lucien2002data} for measuring spectral distortion, and spatial correlation coefficient (SCC) \cite{zhou1998wavelet} for measuring spatial distortion. Lower is better for ERGAS, whereas higher is better for SCC. Note that in the \textit{original scale} scenario, there are no ground truth PS images for comparison. Therefore, in this letter, given a PS output at the original scale, SCC$_0$ between the network output and PAN input, ERGAS$_1$ between a down-scaled network output and its MS input, and SCC$_1$ between the down-scaled network output and down-scaled PAN input were computed. \par
Table \ref{table:1} shows average quality metric scores (with standard errors), ERGAS$_1$, SCC$_1$ and SCC$_0$ for PS results at the PAN resolution (0.3 $m$ GSD). Here, 100 MS-PAN pairs from the WorldView-3 satellite test dataset were selected for testing in the \textit{original scale} scenario. As shown in Table \ref{table:1}, the PS results by PanNet, BDPN and DSen2 have lower ERGAS values, showing lower spectral distortion, but lower SCC values, indicating higher spatial distortion. On the other hand, our methods with the S3 loss generated the PS images with much higher SCC values, but with slightly higher spectral distortion. Note that, since ERGAS simply computes the score values of spectral distortion between MS and PAN test image pairs that are often misaligned with unknown magnitudes and directions, it may not be effective in measuring the distortions for misaligned MS-PAN pairs. \par
Here, we additionally propose a more effective spectral distortion metric, called \textit{n-ERGAS}, which is a simple variant of ERGAS inspired by an evaluation method used in the NTIRE 2018 Super-Resolution Challenge \cite{timofte2018ntiresr} for misaligned input-target pairs. In this challenge, input images were randomly translated from the corresponding target images, and a new metric was used for evaluation. As for our n-ERGAS, once we obtain a PS result image, $\pm$6-pixel translations are applied to obtain 144 translated PS images. Next, multiple ERGAS scores are computed using down-scaled versions of these translated images and an MS input, and the most favorable (smallest) ERGAS score is selected as the final ERGAS score for evaluation. As methods using our proposed S3 loss can reconstruct spectral information of misaligned MS on spatially correlated areas with PAN, their n-ERGAS scores should be lower (thus better) than the corresponding ERGAS scores. The n-ERGAS$_1$ scores for the various methods are presented in Table \ref{table:1}. As shown, all our methods (PanNet-S3, BDPN-S3 and DSen2-S3) have lower n-ERGAS scores compared to the corresponding ERGAS scores, while almost no difference is observed between n-ERGAS and ERGAS scores for the baselines (PanNet \cite{Yang2017}, BDPN \cite{zhang2019pan} and DSen2 \cite{Lanaras2018}). This indicates that our S3 loss indeed tries to minimize the spectral distortion more on spatially correlated areas with PAN, demonstrating the effectiveness of using our S3 loss for misaligned MS-PAN images.
\par We now visually compare several pan-sharpening methods including ours. Fig. \ref{fig:2} shows PS images for various methods on WV3. First, PS images provided from the dataset show high spectral distortion, with blue glow around cars. Since trained using a simple loss between network outputs and MS targets, PanNet, BDPN and DSen2 tend to perform poorly on misaligned MS-PAN test inputs, creating unpleasant artifacts around strong edges and moving objects in the PS images. On the other hand, our method using the proposed S3 loss can reconstruct PS images with highly sharpened edges, rooftops, roads and cars with much less artifacts, visually outperforming the conventional methods. However, some spectral artifacts are slightly visible around cars, indicating that there is still room for improvement. Nevertheless, the results using conventional PanNet, BDPN and DSen2 methods still suffer from ghosting and double-edge artifacts, degrading the overall visual quality. This confirms that our proposed S3 loss can be used for various networks to generate PS images with higher visual quality and less artifacts, compared to their baselines. \par
Moreover, we conducted experiments using two additional satellite datasets: the WorldView-2 (WV2) dataset and the KOMPSAT-3A (K3A) dataset. The WV2 dataset is of 11 bits per pixel, and includes PAN images of 0.5 m GSD and MS images of 2.0 m GSD. The K3A dataset is of 14 bits per pixel, and includes PAN images of 0.7 m GSD and MS images of 2.8 m GSD. Fig. \ref{fig:3} and \ref{fig:4} show pan-sharpening results at the original scale using various methods on the WV2 dataset and the K3A dataset, respectively. As shown, similar to the experiment results using the WV3 dataset, PS images using our DSen2-S3 method trained with WV2 have a slightly higher spectral distortion compared to MS inputs (higher ERGAS), but their spatial details are much similar to those of PAN inputs (higher SCC). This implies that our S3 loss is effective and robust for different types of satellite datasets. \par
We now present experiment results to show the effectiveness of using the correlation map in our S3 loss. Here, we set $\mathbf{S}=\mathbf{1}$, so that the correlation map was not used in training. Fig. \ref{fig:5} shows pan-sharpening results at the original scale on the WorldView-3 test dataset for our DSen2-S3 without using the correlation maps $\mathbf{S}$. As shown, simply adding the spatial loss regarding PAN inputs would not be able to overcome artifacts, and much more spectral distortions are visible around moving cars if we do not incorporate the correlation map into our S3 loss. Therefore, we can confirm that the correlation map plays an important role in our S3 loss.
%More extensive experiment results are included in our Supplementary Material\footnote{link}.

\section{Conclusion}
We proposed a novel spectral-spatial structure (S3) loss that can be effectively applied for CNN-based pan-sharpening methods. Our S3 loss is featured with a combined measuring capability of spectral, spatial and structural distortions, so that the CNN-based pan-sharpening networks can be effectively trained to generate highly detailed PS images with less artifacts, compared to the conventional losses simply based on the difference between network outputs and MS targets.

%\clearpage
\bibliographystyle{IEEEtran}  % -J
\bibliography{mybib_190803}  % -J

\end{document}